# Feature Selection based on PCA and PSO for Multimodal Medical Image Fusion using DTCWT

Padmavathi K[a,*], Mahima Bhat[a] and Maya V Karki[b]

[a,*]Department of Electronics and Communication Engineering, NMAM Institute of Technology, Nitte, 574110, Karnataka, India
[a]Department of Electronics and Communication Engineering, NMAM Institute of Technology, Nitte, 574110, Karnataka, India
[b]Department of Electronics and Communication Engineering, MSR Institute of Technology, Bengaluru, 560054, Karnataka, India

[a,*]padmavathik@nitte.edu.in [a]bhatmahi92@gmail.com [b]mayavkarki@msrit.edu

**Abstract**
Multimodal medical image fusion helps to increase efficiency in medical diagnosis. This paper presents multimodal medical image fusion by selecting relevant features using Principle Component Analysis (PCA) and Particle Swarm Optimization techniques (PSO). DTCWT is used for decomposition of the images into low and high frequency coefficients. Fusion rules such as combination of minimum, maximum and simple averaging are applied to approximate and detailed coefficients. The fused image is reconstructed by inverse DTCWT. Performance metrics are evaluated and it shows that DTCWT-PCA performs better than DTCWT-PSO in terms of Structural Similarity Index Measure (SSIM) and Cross Correlation (CC). Computation time and feature vector size is reduced in DTCWT-PCA compared to DTCWT-PSO for feature selection which proves robustness and storage capacity.

## 1. Introduction

Nowadays, acquiring high resolution and more informative description of human anatomies and functions becomes possible due to the rapid advances in medical imaging technologies. Such development encourages the research in medical image analysis field. Multimodal medical images play a very important role in medical diagnostics. Various medical images exist with each having unique characteristics. Fusion of these images helps in accurate diagnostics and also it can be used in e-health care. Image fusion techniques have drawn attention in combining and enhancing information. The aim of the image fusion technique is to obtain a more detailed and informative resulting image. The ultimate usefulness of image fusion is the quality of the information contained in the output images and representing the information more compactly (El-Gamal et al. 2016; Raol 2010) .

Imaging modality such as Computer Tomography (CT) has high spatial resolution and geometrical characteristics which clearly displays the bony structure. Magnetic Resonance Imaging (MRI) displays the soft tissues and organs. Hence the combination of CT and MRI images provides more information with regard to pathological conditions of relevant organs and increases the diagnostic capabilities in clinical applications (Pappachen & Dasarathy 2014).

In recent years, many software solutions have been developed to quantify the image fusion problems. Fusion schemes based on transform domain methods such as DWT, DTCWT and DTCWT-PCA fusion has attracted the researcher's attention. DTCWT-PCA based fusion scheme has given best results in terms of phase and directional information, reduced storage and computation time.
In this work, DTCWT was applied to the input images to obtain approximate and detailed coefficients. Important features were selected and optimized by applying principle component analysis and particle swarm optimization techniques from the decomposed coefficients. To these coefficients, combinations of basic fusion rules are applied to generate optimal results.

## 2. Implementation Details

### 2.1 Dual Tree Complex Wavelet Transform

DWT provides a compact representation of frequency content present in the image but does not provide sufficient directional information and results in an image with shift invariance and additive noise. A recent advancement to DWT with important additional properties is dual tree complex wavelet transform. DTCWT is an over complete wavelet transform and provides sparse representation and useful in characterization of structure of an image. It is nearly shift invariant directional selective and computationally effective over DWT. It also preserves time frequency information and hence is a suitable approach for medical image fusion. It provides increased memory usage and reduced computation time. It is also able to distinguish between six orientations, which define directions at each level of decompositions (Selesnick et al. 2005). DTCWT gives a perfect reconstruction. The main difference between DWT and DTCWT is that it uses two filter trees instead of one. The improved directional selectivity of DTCWT is important in order to properly reflect the content of the images across edges, boundaries and other important directional features. It is applied to images by separable complex filtering in 2D. The use of DTCWT for image fusion hence gives a drastic quantitative and qualitative improvement over the real valued wavelet transform (Selesnick et al. 2005). DTCWT with its directional capabilities preserves edge information which is very much essential in medical image fusion.

Fig. 1 shows the tree structure of DTCWT with analysis filter banks and Fig. 2 shows orientations present in DTCWT. DTCWT produces six subbands at each scale for both real and imaginary parts at $\pm 15^0$, $\pm 45^0$, $\pm 75^0$ which proves the improvement in directional selectivity which is not achieved by normal DWT (Selesnick et al. 2005).

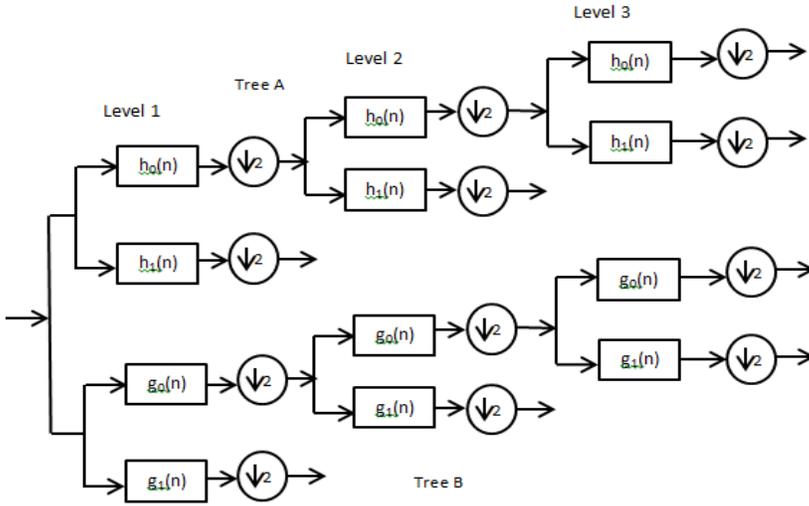 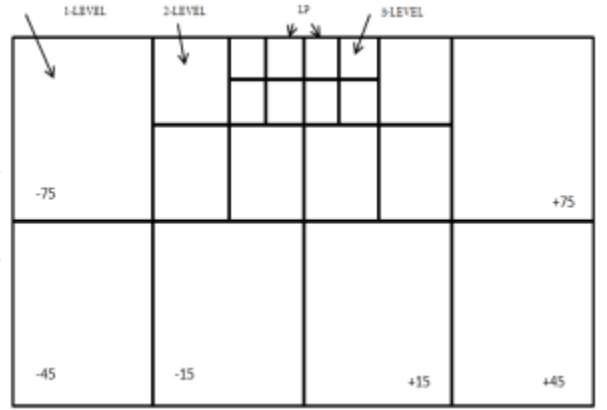

Fig. 1 The tree structure of DTCWT with analysis bank (Selesnick et al. 2005)      Fig. 2 Orientations present in DTCWT

## 2.2 PCA Algorithm

As medical images are bulky, to reduce the data, PCA method is essential. PCA removes the redundant information present in the image (P 2015). It is the simplest of the true eigen vector based multivariate analysis which represents variance in the data. This method determines the weights for each source images using the eigen vector corresponding to the largest eigen value of the covariance matrix of each source image. These are arranged in descending order. The column vector corresponding to larger eigen value is normalized by dividing each element with mean of eigen vector as $P_1 = \frac{V(1)}{\Sigma V}$ and $P_2 = \frac{V(2)}{\Sigma V}$. These normalized eigen vector values act as weight values that are multiplied with each pixel of the DTCWT decomposed input images.

## 2.3 PSO Algorithm

Particle Swarm optimization is a population based optimization algorithm. It is one of the most popular evolutionary computational techniques. In an optimization problem, a function is to be maximized or minimized. The function that has to be optimized is known as objective function or performance index (Arora 2015; April et al. 2015; Patil & Deshpande 2015; Palupi Rini et al. 2011; P 2015; Nahvi & Mittal 2014).

In PSO process, the steps are described as follows:

*Step1:* Initialize $i_{max}, w_1, \emptyset_1, \emptyset_2, n\ (population\ size), x_{i,min}, x_{i,max}$

*Step2:* Initialize starting position and velocities of the variables as, $x_{i,k} = x_{i,min} + (x_{i,max} - x_{i,min})\mu_i,\ v_{i,k} = 0$

*Step 3:* Compute $p_{i,k} = f(x_{i,k}), k = 1,2,3 \ldots n$

*Step4:* Compute $p_{best i,k} = p_{i,k}$ and $g_{best\ i} = minimum\ (p_{best\ i,k})$

*Step5:* Update velocity as $v_{i+1,k} = w_1 v_{i,k} + \emptyset_1 (p_{x_{i,k}} - x_{i,k})\mu_i + \emptyset_2 (g_{ix} - x_{i,k})\mu_i$

*Step 6:* Update position as $x_{i+1,k} = x_{i,k} + v_{i+1,k}$

*Step 7:* Update the fitness function as $p_{i+1,k} = f(x_{i+1,k})$

*Step 8:* If $p_{i+1,k} < p_{best\ i,k}$ then, $p_{best\ i+1,k} = p_{i+1,k}$

*Step 9:* Update $g_{best\ i+1} = minimum\ (p_{best\ i+1,k})$

*Step 10:* If $i < i_{max}$, the increment $i$ and go to *Step 5*, else *Stop*.

The update process is repeated until the maximum number of generation is reached or specified fitness function is achieved.

## 2.4 Fusion Rules

There exists mainly three fusion rules and they are pixel level, feature level and decision level. Usually, because of easy implementation and less computational time, pixel level fusion is employed for medical image fusion. But in the proposed method feature level fusion has been employed by considering spatial domain fusion methods.

Spatial domain fusion methods directly deal with pixels of the input images (Indira et al. 2015). Some of them are

a) *Simple Averaging rule:* In this method, the relevant fused image is obtained by taking the average intensity of corresponding pixels from both the input images.

$$I_f(i,j) = \frac{[I_1(i,j) + I_2(i,j)]}{2} \qquad (1)$$

b) *Maximum Selection rule:* In this method, the fused image can be obtained by selecting the maximum intensity of corresponding pixels from both the input images.

$$I_f(i,j) = \sum_{i=0}^{m}\sum_{i=0}^{n} max[I_1(i,j)I_2(i,j)] \quad (2)$$

c) *Minimum selection rule:* In this technique, the resultant fused image is obtained by selecting minimum intensity of corresponding pixels from both the input images.

$$I_f(i,j) = \sum_{i=0}^{m}\sum_{i=0}^{n} min[I_1(i,j)I_2(i,j)] \quad (3)$$

where $I_1(i,j)$ and $I_2(i,j)$ are the two input images, $I_f(i,j)$ is the fused image.

**3. Proposed Methodology**

3.1 Description

Medical images have larger dimension and are rich in information content. Block diagram of the proposed fusion scheme is shown in Fig. 3. In this work, the source images considered are the combination of MR and CT images of brain. These RGB images are converted to gray scale. Feature selection involves reducing the high dimension data to a lower fewer dimensions. The advantage is that it becomes easy to visualize the data. It also reduces the computation time and storage space required.

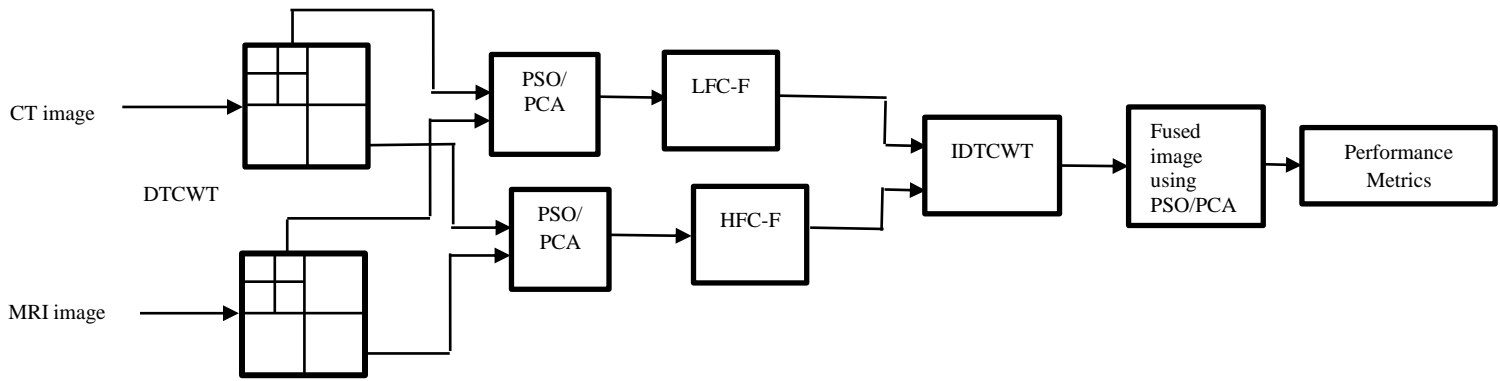

Fig. 3 Proposed Fusion Scheme with PSO/PCA for Feature Selection using DTCWT

3.2 Implementation Steps

In Fig.3, LFC represents the low frequency coefficients and HFC represents the high frequency coefficients.

*Step 1* Two input images CT and MRI that has to be fused are read.
*Step 2* CT and MR images were decomposed into low frequency component LFC-CT, LFC-MR and high frequency component HFC-CT, HFC-MR by using DTCWT. LFC consists of LL coefficients and HFC consists of LH, HL and HH coefficients.
*Step 3* To reduce feature dimension and optimize the features PCA and PSO were applied on wavelet coefficients.
*Step 4* The processed coefficients were fused based on different combinations of fusion rules applied separately to low frequency coefficients LFC-F and high frequency coefficients HFC-F.
*Step 5* The fused LFC and HFCs were combined and inverse DTCWT was applied to obtain fused image.
*Step 6* The final fused image was displayed and the performance metrics were evaluated.

When performing analysis of the complex data, one of the major problems identified is the number of variables involved. PSO is particularly suitable for optimizing a large number of parameter values efficiently. This provides a great potential to edge and corner detection where many pixel positions need to be found. PSO technique has been greatly applied to select and extract important features of an image like its edges, corners and textured regions. Literatures show as how PSO is applied to fused images for feature extraction. In this work, PSO is applied to the decomposed images to select the relevant features for image fusion. This technique of applying PSO for feature selection directly on decomposed images performs better compared to extracting features after fusion.

PCA is a dimensionality reduction technique that is often used to transform a high dimensional dataset into a smaller dimensional subspace. Reducing the dimensionality via PCA can simplify the dataset that facilitates description, visualization and insight. In this work PCA was applied to the decomposed coefficients. It removes the redundant information present in DTCWT. PCA is used to highlight the internal structures of the images that explains the variance in the images.

## 4. Performance Evaluation Metrics

The main aim of an image fusion process is that all functional and efficient information must be secured. At the same time, the reconstructed image must not be changed due to the undesirable introduction of artifacts (Kaur & Sharma 2016).
Performance evaluation of the proposed method is conducted using various parameters. No reference methods such as Entropy (EN), Standard Deviation (SD), Structural Similarity Index Measure (SSIM), and Cross Correlation (CC) are responsible for the restored information content in the fused image. Objective evaluation of the fused image quality can be obtained by using full reference methods such as Peak signal to Noise Ratio (PSNR).

*1) Entropy (E):* Entropy is used to measure the information content of a fused image. Higher the entropy value, richer is the information content in the fused image. It is given by
$E = \sum_{i=0}^{L} h_{I_f}(i) \log_2 h_{I_f}(i)$. $'E'$ is the entropy of fused image and $'h_{I_f}'$ is the histogram count of the fused image.

*2) Standard Deviation (SD):* It is used to measure contrast in the fused image. An image with high contrast would have a high standard deviation. It is given
by $\sigma = \sqrt{\sum_{i=0}^{L}(i - \bar{i})^2 h_{I_f}(i)}$, where $'\sigma'$ is the standard deviation of the fused image, $'h_{I_f}'$ is the histogram counts of the fused image, $'i'$ is the index of summation and $'\bar{i}'$ is mean of histogram.

*3) Structural Similarity Index Measure (SSIM):* It is used to compare the local patterns of pixel intensities between the reference and fused images. The range varies between -1 and +1. The value +1 indicates the reference and fused images are similar.
It is given by $SSIM = \frac{(2\mu_{I_r}\mu_{I_f} + C_1)(2\sigma_{I_r I_f} + C_2)}{(\mu^2_{I_r} + \mu^2_{I_f} + C_1)(\sigma^2_{I_r} + \sigma^2_{I_f} + C_2)}$, where $'\mu'$ is the average value of the fused image, $I_r$ and $I_f$ are the reference and the fused images, $C_1$ and $C_2$ are constants, with $C_1 = (0.01 * L)^2$ and $C_2 = (0.03 * L)^2$ and $'L'$ is the specified range.

*4) Cross Correlation (CC):* It measures the degree of correlation between the fused image and the reference image. It is used to measure the similarity of spectral features between the two images. The value of CC must be close to +1 which indicates that the reference image and fused image are same. It is given by $CC = \frac{2C_{rf}}{C_r + C_f}$, where $C_r$ and $C_f$ correlation coefficients of reference and fused image.

*5) Peak Signal to Noise Ratio (PSNR)* It is a widely used metric. It is computed as the number of gray levels in the image divided by the corresponding pixels in the reference and fused images. A higher value of PSNR indicates superior fusion. It is given by
$PSNR = 20 \log_{10} \left[ \frac{L^2}{\frac{1}{MN} \sum_{i=1}^{M} \sum_{j=1}^{N} (I_r(i,j) - I_f(i,j))^2} \right]$, where $I_r(i,j)$ is the reference image and $I_f(i,j)$ is the fused image.

## 5. Results and Discussions

In this work, feature selection is performed using PCA and PSO algorithms. In order to evaluate the effectiveness and validity of the techniques, two sets of medical images are chosen to conduct experiments. The images are of size 512*512. The medical images are obtained from the database "The Whole Brain Atlas", of Harvard University. Figure 4(a)-(b) shows the CT and MRI images of a normal brain considered as data set 1. Figure 4(c)-(d) shows the CT and MRI images of a brain with tumorous region considered as data set 2. Experiments are carried out using MATLAB R2014a simulation tool with Intel® Core™ i7-5500U CPU @2.4GHz and 4GB RAM.

In this experiment, DTCWT based medical image fusion is carried out with decomposition level as 2 so as to get better directionality, shift invariance and phase information. Feature selection is carried out using two techniques ie., PCA and PSO. Both methods are compared with qualitative measure such as contrast, information content with minimum error and signal to noise ratio. Various combinations of fusion rules such as Avg-Avg, Avg-Max, Max-Avg, Max-Max, Min-Avg and Min-Max are applied to HFC and to the LL band. DTCWT-PSO based method of feature selection gives best details with respect to corners and edges. But there are discontinuities in curvatures as shown in Fig. 5(a1)-(a6) and 5(c1)-(c6) . The DTCWT-PCA based method eliminates distortion and discontinuities seen in case of DTCWT-PSO and also gives high textural contents with improved contrast as shown in Fig. 5(b1)-(b6) and 5(d1)-(d6). In PSO  each particle updates its velocity through a linear combination among its present status, but becomes inefficient when it searches in a complex space. As DTCWT give complex space, DTCWT with PSO  is less efficient compared to DTCWT with PCA. The feature vector size of the fused image is depicted in Table 1. Compared to DTCWT-PSO, DTCWT-PCA gives better performance with reduced feature vector size and and higher computational efficiency. The statistical parameters such as EN, SD, SSIM, CC and PSNR are high for DTCWT-PCA compared to DTCWT-PSO as shown in Table 2 and Table 3. The corresponding graphs are shown in Fig. 6(a)-6(e). This indicates that feature selection with dimensionality reduction performs better compared to feature selection with optimization. Fusion rules such as Avg-Max gives high information content and high contrast in data set 1. Max-Avg rule gives high SSIM, CC and PSNR. Hence from a subjective point of view, the proposed method DTCWT-PCA gives better results and the fused image is more clear and accurate.

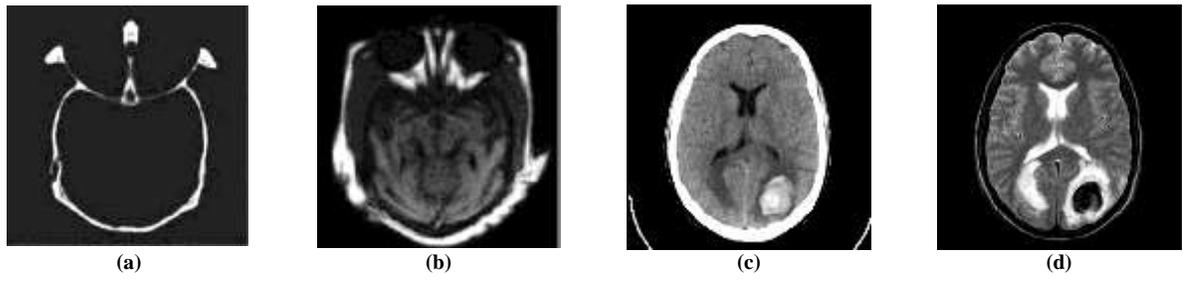

| (a) | (b) | (c) | (d) |

Fig. 4(a)-(b) Data Set 1: CT and MRI of a Normal Brain 4(c)-(d) Data Set 2: CT and MRI with tumor

| **FUSION RULES** | **DATA SET 1** | | **DATA SET 2** | |
|---|---|---|---|---|
| | **PSO** | **PCA** | **PSO** | **PCA** |
| **AVG-AVG** | 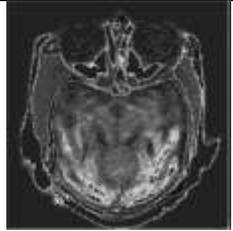 (a1) | 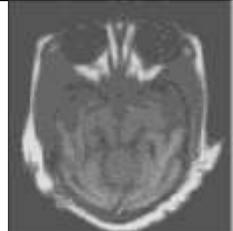 (b1) | 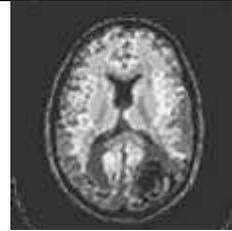 (c1) | 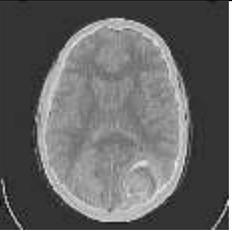 (d1) |
| **AVG-MAX** | 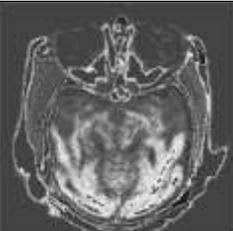 (a2) | 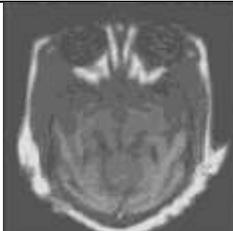 (b2) | 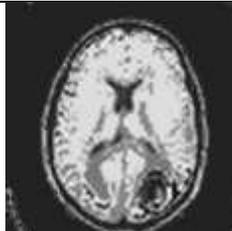 (c2) | 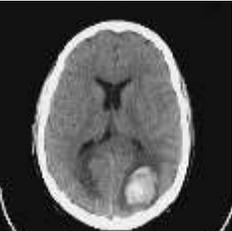 (d2) |
| **MAX - AVG** | 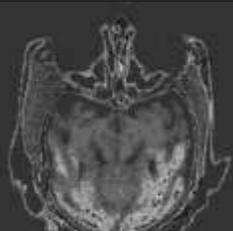 (a3) | 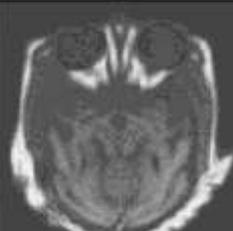 (b3) | 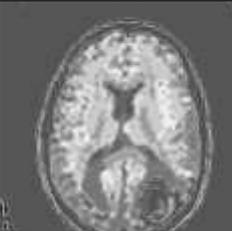 (c3) | 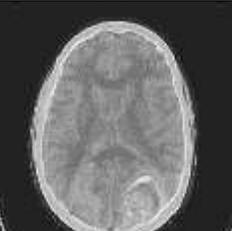 (d3) |
| **MAX -MAX** | 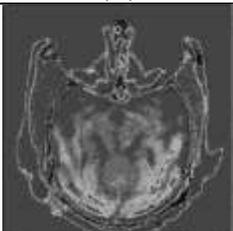 (a4) | 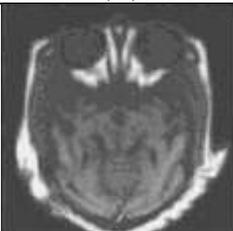 (b4) | 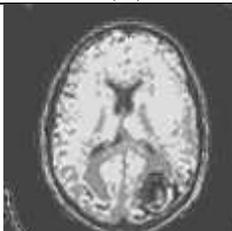 (c4) | 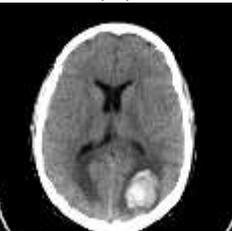 (d4) |
| **MIN - AVG** | 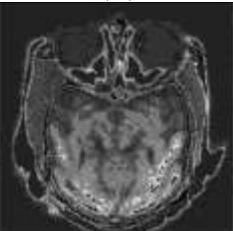 (a5) | 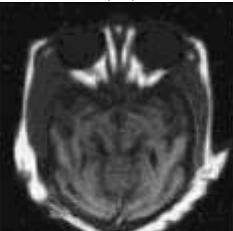 (b5) | 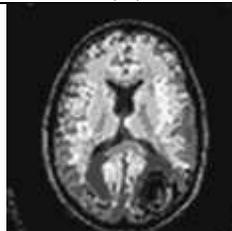 (c5) | 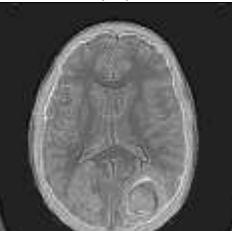 (d5) |
| **MIN -MAX** | 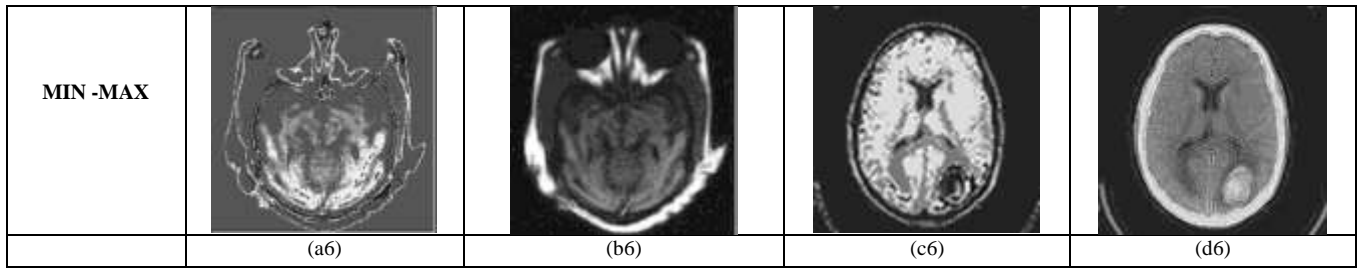 | | | |
| | (a6) | (b6) | (c6) | (d6) |

Fig. 5 Fusion results of Data Set 1 (a1)-(a6), (b1-b6) with PSO and PCA for feature selection and dimensionality reduction with various combinations of fusion rules. Fusion results of Data Set 2 (c1-c6), (d1-d6) with PSO and PCA for feature selection and dimensionality reduction with various combinations of fusion rules.

**Table 1**
Feature Vector Size

| | Data set 1 | | Data set 2 | |
|---|---|---|---|---|
| **Fusion Rules** | **PSO** | **PCA** | **PSO** | **PCA** |
| **AVG-AVG** | 19kB | **9kB** | 12kB | **10kB** |
| **AVG-MAX** | 11kB | **10kB** | 12kB | **9kB** |
| **MAX - AVG** | 22kB | **7kB** | 13kB | **9kB** |
| **MAX -MAX** | 21kB | **8kB** | 13kB | **6kB** |
| **MIN - AVG** | 19kB | **6kB** | 11kB | **9kB** |
| **MIN -MAX** | 11kB | **7kB** | 12kB | **10kB** |

**Table 2**
Performance Evaluation Indices Of Data Set 1

| Fusion Rules | PSO | | | | | PCA | | | | |
|---|---|---|---|---|---|---|---|---|---|---|
| | **EN** | **SD** | **SSIM** | **CC** | **PSNR** | **EN** | **SD** | **SSIM** | **CC** | **PSNR** |
| **AVG-AVG** | 4.7494 | 12.2487 | 0.3957 | 0.2026 | 14.1832 | 6.8714 | 57.0215 | 0.6994 | 0.7491 | 16.3423 |
| **AVG-MAX** | 5.0847 | 19.4579 | 0.6529 | 0.1858 | 15.7229 | **6.9055** | **59.5099** | 0.6683 | 0.7270 | 15.6604 |
| **MAX - AVG** | 4.6402 | 12.3207 | 0.4372 | 0.2042 | 14.2945 | 6.8499 | 56.2623 | **0.7236** | **0.7534** | **16.5306** |
| **MAX -MAX** | 4.7978 | 14.6678 | 0.6999 | 0.1556 | 16.0882 | 6.8931 | 58.8049 | 0.6889 | 0.7302 | 15.8156 |
| **MIN - AVG** | 4.7078 | 11.9730 | 0.4844 | 0.1540 | 14.4173 | 6.7961 | 55.9846 | 0.7094 | 0.7340 | 16.3616 |
| **MIN -MAX** | 4.8906 | 19.3576 | 0.6653 | 0.1285 | 15.8254 | 6.8224 | 58.5905 | 0.6760 | 0.7109 | 15.6601 |

**Table 3**
Performance Evaluation Indices Of Data Set 2

| Fusion rules | PSO | | | | | PCA | | | | |
|---|---|---|---|---|---|---|---|---|---|---|
| | **EN** | **SD** | **SSIM** | **CC** | **PSNR** | **EN** | **SD** | **SSIM** | **CC** | **PSNR** |
| **AVG-AVG** | 4.3270 | 29.5495 | 0.5507 | 0.6274 | 12.6930 | 4.9034 | 66.9173 | 0.7579 | 0.9412 | 20.8015 |
| **AVG-MAX** | 4.3744 | 39.3385 | 0.5825 | 0.7200 | 14.3396 | 4.7723 | 82.5462 | 0.6463 | 0.8179 | 14.2430 |
| **MAX - AVG** | 4.3033 | 29.9203 | 0.5570 | 0.6314 | 12.7434 | 4.8577 | 66.5632 | 0.7904 | 0.9470 | 21.2697 |
| **MAX -MAX** | 4.4037 | 39.1627 | 0.5915 | 0.7221 | 14.3335 | 4.6406 | **82.5672** | 0.6637 | 0.8199 | 14.2778 |
| **MIN - AVG** | 4.3265 | 29.6851 | 0.5392 | 0.6187 | 12.6677 | 4.9645 | 65.3258 | **0.8684** | **0.9596** | **22.5159** |
| **MIN -MAX** | 4.4081 | 39.1955 | 0.5654 | 0.7123 | 14.2508 | **5.0411** | 79.4195 | 0.7216 | 0.8344 | 14.9281 |

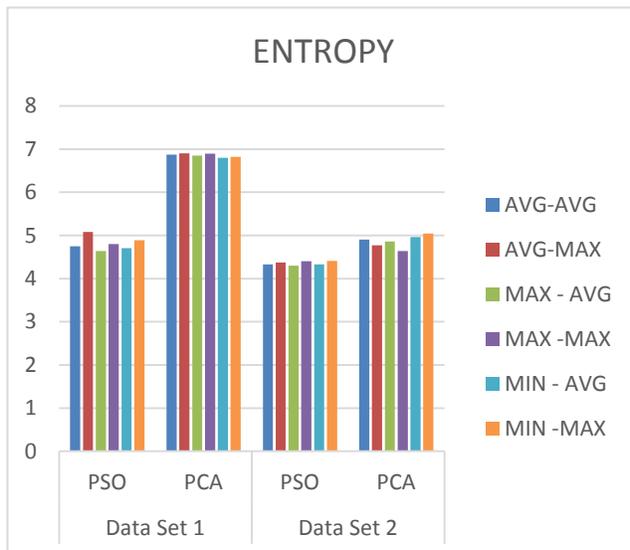

Fig. 6 (a) Entropy plot of fused image with PSO/PCA and Fusion rules

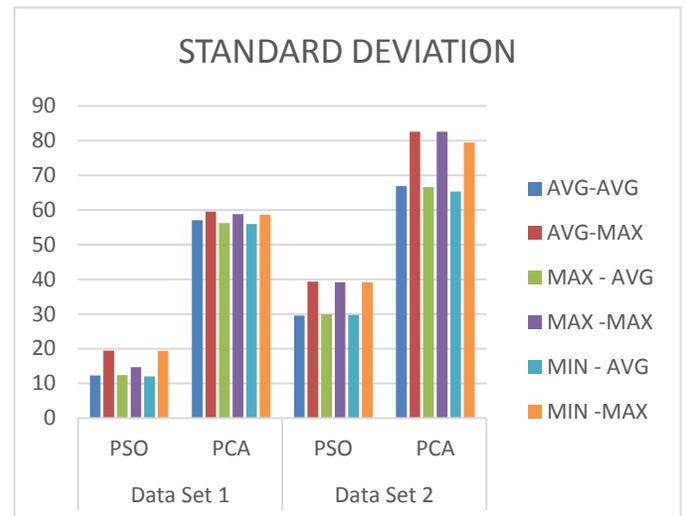

Fig. 6 (b) Standard deviation plot of fused image with PSO/PCA and fusion rules

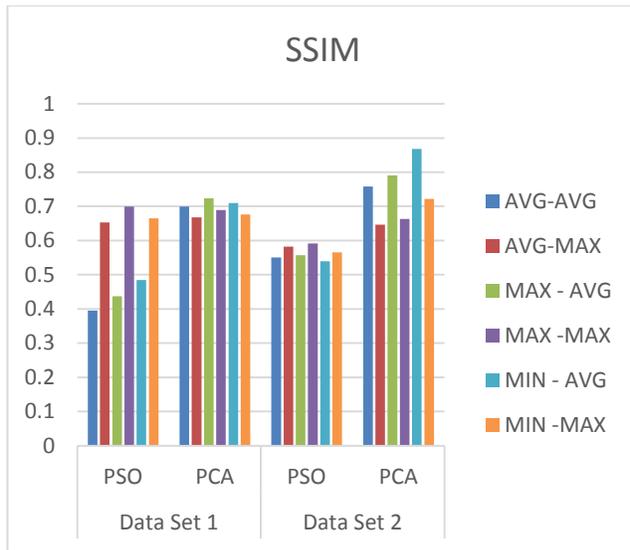
Fig. 6 (c) SSIM plot of fused image with PSO/PCA and fusion rules

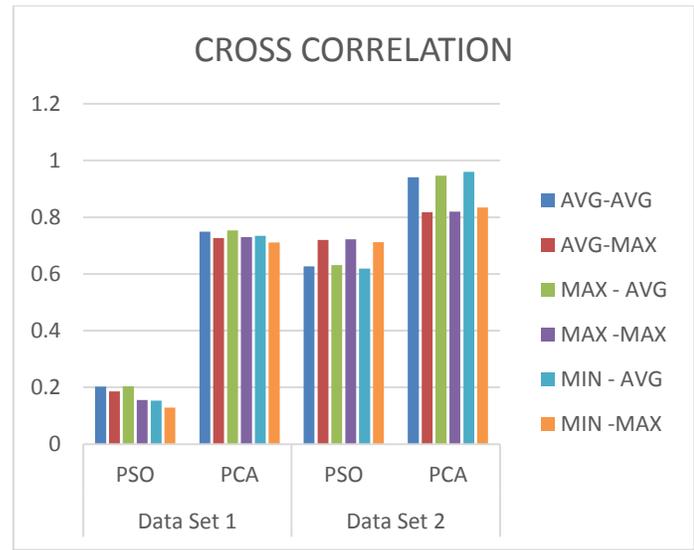
Fig. 6 (d) Cross Correlation plot of fused image with PSO/PCA and fusion rules

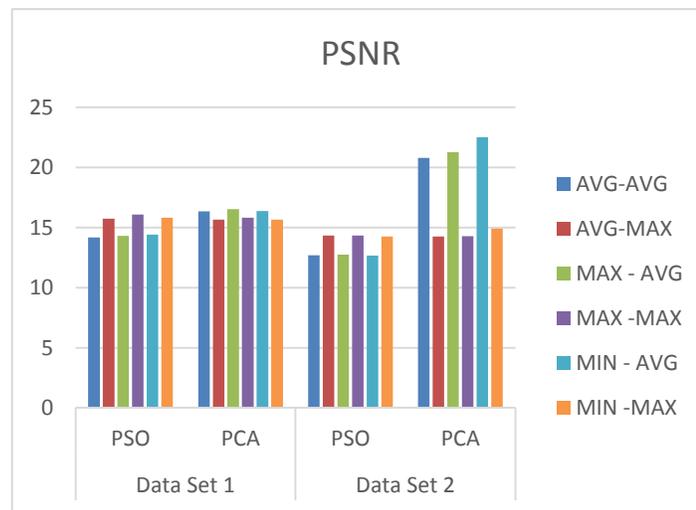
Fig. 6 (e) PSNR plot of fused image with PSO/PCA and fusion rules

## 6. Conclusion

In this paper, multimodal medical image fusion using DTCWT-PCA and DTCWT-PSO for feature selection is presented. Compared with DTCWT-PSO, DTCWT-PCA gives better performance with reduced feature vector size and high computational efficiency. Compared with traditional fusion rules, the proposed fusion rules with various combinations gives best results. Both visual and objective analysis show that DTCWT-PCA method suits better for medical image fusion and feature selection.

**Acknowledgement**

**References**

April, I., Kumar, R.E.S. & Rajesh, T., 2015. AN IMPROVED PSO BASED COEFFICIENT SELECTION FOR MEDICAL IMAGE FUSION. *International Research Journal of Emerging Trends in Multidisciplinary*, 1(2), pp.17–24. Available at: www.irjetm.com.

Arora, R.K., 2015. *Optimization: Algorithms and Applications*, CRC Press, Taylor and Francis group. Available at: http://www.amazon.com/Optimization-Applications-Rajesh-Kumar-Arora.

El-Gamal, F.E.-Z.A., Elmogy, M. & Atwan, A., 2016. Current trends in medical image registration and fusion. *Egyptian Informatics Journal*, 17(1), pp.99–124. Available at: http://dx.doi.org/10.1016/j.eij.2015.09.002%5.

Indira, K.P., Rani Hemamalini, R. & Indhumathi, R., 2015. Pixel based medical image fusion techniques using discrete wavelet transform and Stationary wavelet transform. *Indian Journal of Science and Technology*, 8(26), pp.1–7.

Kaur, A. & Sharma, R., 2016. Stationary Wavelet Transform Image Fusion and Optimization Using Particle Swarm Optimization. *IOSR Journal of Computer Engineering (IOSR-JCE)*, 18(3), pp.32–38. Available at: www.iosrjournals.org.

Nahvi, N. & Mittal, D., 2014. Medical Image Fusion Using Discrete Wavelet Transform. *Journal of Engineering Research and Applications www.ijera.com*, 4(5), pp.165–170. Available at: www.ijera.com.

P, A.M.N., 2015. Comparative Analysis of Transform Based Image Fusion Techniques for Medical Applications. In *Innovations in Information, Embedded and Communication Systems (ICIIECS)*. pp. 0–5.

Palupi Rini, D., Mariyam Shamsuddin, S. & Sophiyati Yuhaniz, S., 2011. Particle Swarm Optimization: Technique, System and Challenges. *International Journal of Computer Applications*, 14(1), pp.19–27. Available at: http://www.ijcaonline.org/archives/volume14/number1/1810-2331.

Pappachen, A. & Dasarathy, B. V, 2014. Medical image fusion : A survey of the state of the art. *Information Fusion*, 19, pp.4–19. Available at: http://dx.doi.org/10.1016/j.inffus.2013.12.002.

Patil, P.P. & Deshpande, K.B., 2015. New Technique for Image Fusion Using DDWT and PSO In Medical field. *International Journal on Recent and Innovation Trends in Computing and Communication*, 3(4), pp.2251–2254. Available at: http://www.ijritcc.org.

Raol, J., 2010. *Multi-sensor data fusion with MATLAB*, CRC Press, Taylor and Francis group. Available at: http://onlinelibrary.wiley.com/doi/10.1002/cbdv.200490137.

Selesnick, I.W., Baraniuk, R.G. & Kingsbury, N.C., 2005. The dual-tree complex wavelet transform. *IEEE Signal Processing Magazine*, 22(6), pp.123–151. Available at: http://ieeexplore.ieee.org/document/1550194/.

http://www.med.harvard.edu/anlib/home.html, "The Whole Brain Atlas Database", Keith A.Johnson, J.Alex Becker, Harvard university.